\documentclass[journal,twoside, a4paper]{IEEEtran}
\IEEEoverridecommandlockouts

\usepackage{color}
\usepackage{amssymb}
\usepackage{rotating}
\usepackage[T1]{fontenc}
\usepackage{lscape}
\usepackage{dblfloatfix}
\usepackage{xcolor}
\usepackage{subcaption}
\usepackage[ruled,vlined]{algorithm2e}
\usepackage[autostyle]{csquotes}
\usepackage{tikz}
\usepackage{acro}
\usepackage{siunitx}
\usepackage{enumitem}
\usepackage{multirow}
\usepackage{cite}
\usepackage{hyperref}
\usepackage{amsmath,amssymb,amsfonts}
\usepackage{tikz}
\usepackage{float}
\usepackage{algorithmic}
\usepackage{graphicx}
\usepackage{textcomp}
\usepackage{xcolor}
\usepackage{url}
\usepackage{tabularx}
 \usepackage{booktabs} 
\usepackage{comment}
\def\BibTeX{{\rm B\kern-.05em{\sc i\kern-.025em b}\kern-.08em
    T\kern-.1667em\lower.7ex\hbox{E}\kern-.125emX}}

\ifCLASSINFOpdf
\else
 
\fi

\begin{document}

\title{Hybrid Whale-Mud-Ring Optimization for Precise Color Skin Cancer Image Segmentation\\
}

\author{\IEEEauthorblockN{Amir Hamza\textsuperscript{1}, Badis Lekouaghet\textsuperscript{2}, and Yassine Himeur\textsuperscript{3}}\\
	\IEEEauthorblockA{\textsuperscript{1}Non-Destructive Laboratory, Faculty of Technology, University of Jijel, 18000, Jijel,
Algeria.\\
		\textsuperscript{2}Research Center in Industrial Technologies (CRTI), Cheraga, Algiers, P.O. Box 64, 16014, Algeria\\
        \textsuperscript{3}College of Engineering and Information Technology, University of Dubai, Dubai, UAE\\
		Emails: \textsuperscript{1}amir.hamza@univ-jijel.dz,
		\textsuperscript{2}b.lekouaghet@crti.dz, \textsuperscript{3}yhimeur@ud.ac.ae}}
\maketitle

\begin{abstract}
Timely identification and treatment of rapidly progressing skin cancers can significantly contribute to the preservation of patients' health and well-being. Dermoscopy, a dependable and accessible tool, plays a pivotal role in the initial stages of skin cancer detection. Consequently, the effective processing of digital dermoscopy images holds significant importance in elevating the accuracy of skin cancer diagnoses. Multilevel thresholding is a key tool in medical imaging that extracts objects within the image to facilitate its analysis. In this paper, an enhanced version of the Mud Ring Algorithm hybridized with the Whale Optimization Algorithm, named WMRA, is proposed. The proposed approach utilizes bubble-net attack and mud ring strategy to overcome stagnation in local optima and obtain optimal thresholds. The experimental results show that WMRA is powerful against a cluster of recent methods in terms of fitness, Peak Signal to Noise Ratio (PSNR), and Mean Square Error (MSE).
\end{abstract}

\begin{IEEEkeywords}
Mud Ring Algorithm, Image segmentation, Skin cancer, Bubble net attack, Cross entropy
\end{IEEEkeywords}

\section{Introduction}
Skin cancer is a malignant tumor that has been responsible for many deaths in recent years. According to the World Health Organization's (WHO) 2020 records, skin cancer was ranked as the fifth most commonly diagnosed cancer~\cite{a1,habchi2023ai}. The American Academy of Dermatology (AAD) further revealed that approximately 9,500 individuals in the United States are diagnosed with skin cancer daily. Their 2022 report also highlighted that at least one in every five Americans will develop skin cancer at some point in their lives. Outside the United States, in regions like Australia and New Zealand, the average incidence rate is 33 cases per 100,000 residents~\cite{a0, a1}. Early detection of melanoma can substantially decrease both mortality rates and the number of people affected by the disease~\cite{a2}. Detecting melanoma in its initial stages can also lead to significant cost savings in treatment. This pressing need for early detection has spurred scientists to devise swift and precise cancer detection techniques, aiming to prolong patients' lives. In this endeavor, segmentation techniques are being explored for their potential role in skin cancer detection. Multilevel thresholding-based image segmentation is a crucial tool in various fields including robotics~\cite{a5}, microscopic image analysis~\cite{a6}, satellite imagery~\cite{a7}, and medical image segmentation \cite{bechar2023harenessing}. It assists medical professionals in understanding complex inner structures of in tricate images, such as those of skin cancer~\cite{a8}, paving the way for accurate diagnoses. These images are often obtained using Dermoscopy, a renowned diagnostic method that reveals detailed skin structures and features non-invasively~\cite{a9,23}.

 Multilevel thresholding image segmentation (MIS) using optimization algorithms is recognized as a vital component in artificial intelligence when addressing healthcare challenges. Within this framework, optimization algorithms are utilized to fine-tune a series of thresholds, striving for an optimal or highly efficient solution.

Unfortunately, metaheuristic algorithms such as the Whale Optimization Algorithm (WOA)\cite{1}, Reptile Search Algorithm (RSA)\cite{2}, Chimp Optimizer (Chimp)\cite{3}, and Mud Ring Algorithm (MRA)\cite{4} face various challenges. These include local optima, premature convergence, and a lack of diversity in the initial population, which lead to subpar segmentation outcomes~\cite{5}. To address these, researchers have devised several solutions with the goal of achieving accurate results and ensuring superior medical diagnostic outcomes. Hamza et al.\cite{6} introduced an effective multilevel thresholding algorithm for COVID-19 diagnosis, utilizing Masi entropy as a fitness function. Sathya et al.\cite{7} explored multilevel thresholding using advantageous maximizing objective functions like Kapur, Otsu, and Minimum Cross Entropy (MCE) to extract precise threshold values for color image segmentation problems. In \cite{8}, the authors proposed a modified firefly algorithm with an opposition-based strategy for multilevel image segmentation, integrating Kapur’s, Tsali’s, and fuzzy entropy-based objective functions to refine the segmentation process. Qi et al.\cite{9} introduced a novel iteration of the Whale Optimization Algorithm (WOA) termed LXMWOA. This iteration features Levy initialization, a directional cross-over step, and a directional mutation operator to counteract sluggish convergence speeds. Hao et al.\cite{10} showcased an enhanced salp swarm algorithm (ILSSA) to mitigate the limitations of meta-heuristic algorithms in multi-threshold image segmentation (MIS). This method was adeptly applied to segment skin cancer dermoscopic images, demonstrating superiority over other techniques. Houssein et al.\cite{11} detailed an efficient Improved Golden Jackal Optimization (IGJO) technique based on Opposition Based Learning (OBL) for medical color image segmentation, with a particular emphasis on skin cancer case studies. Liu et al.\cite{12} researched a new hybrid method, HCROA, which amalgamates Chimp Optimization and Remora Optimization Algorithm (ROA). Designed to combat premature convergence and to explore an expansive search space, it was subsequently applied to medical image segmentation.

While many of the methods previously discussed, along with other algorithms, have shown commendable performance in image segmentation challenges, there remains an imperative need to develop innovative optimization algorithms \cite{46}. This is because no single optimization algorithm exists that can universally address every type of optimization problem \cite{Adam2019, Lekouaghet2022}. Stemming from this need, this paper proposes a new method termed the Whale Optimization Algorithm (WOA) hybridized with the Mud Ring Algorithm (MRA), referred to as WMRA. This method targets multilevel thresholding for color skin cancer medical image segmentation, employing the Minimum Cross Entropy (MCE) as its objective function. The innovative technique draws its inspiration from the behavior of dolphins that encircle their prey in circular patterns. By melding the bubble net attack strategy of the WOA algorithm, which adopts a spiral path, with the chasing mechanism of the MRA, the WMRA method is enhanced in robustness. One significant benefit of this approach is its capability to ensure commendable convergence by honing in on the search space, which assists in evading local optima and leads to superior segmentation performance. The efficacy of the proposed method is evaluated using color images of skin cancer, examining metrics such as the fitness of the red, green, and blue channels, Peak Signal to Noise Ratio (PSNR), and Mean Square Error (MSE). The findings indicate that the introduced WMRA technique offers promising outcomes when juxtaposed with its counterparts, specifically WOA, RSA, Chimp, and MRA. In summary, the primary contributions of this article are outlined as follows:

\begin{itemize}
\item The WMRA presents an enhanced version of the original MRA by integrating the bubble net attack technique from the WOA method, specifically tailored to tackle image segmentation challenges via multilevel thresholding.
\item The application of WMRA is centered on addressing color skin cancer, leveraging the Minimum Cross Entropy (MCE) as the fitness function.
\item A comparative analysis of the proposed method against recent optimization algorithms is provided.
\item The efficacy of the proposed method is evaluated using various metrics, including the fitness of the red, green, and blue channels, Peak Signal to Noise Ratio (PSNR), and Mean Square Error (MSE).
\end{itemize}
 

\section{Materials and Methods}\label{sec:materials}
\subsection{Mud Ring Algorithm (MRA)}
The Mud Ring Algorithm (MRA) emerges as a contemporary and potent methodology, aptly reflecting the intricate behaviors exhibited by dolphins in their quest for sustenance. This cutting-edge approach is divided into two core stages: Exploration (foraging) and Exploitation (feeding). The subsequent subsections provide a detailed mathematical rendition of these pivotal stages.

\subsubsection{Exploration phase}
In this phase, dolphins adopt a variety of tactics to scout for food, capitalizing on their innate chasing patterns. Notably, they engage in a vocalization process, where they produce a sequence of sounds to pinpoint their prey. These sounds have a dynamic range, adapting responsively to the location of the prey. This behavior can be mathematically articulated as depicted in Eq.~(\ref{eq:loudness}).
\begin{equation}\label{eq:loudness}
	\vec{K'}=2 \times \vec{a'} \cdot \vec{r'}-\vec{a'}
\end{equation}

Where, $\vec{r'}$ is a random number in range [0, 1], and $\vec{a'}$ is modeled as follows:

\begin{equation}\label{eq:a}
	\vec{a'}=2\left(1-\frac{\mathrm{it}}{\mathrm{Max} \_itr }\right)
\end{equation}

It is worth noting that the exploration phase is carried out based on the value of the vector $|\vec{K'}|\geq 1$. This guides the global search based on the dolphins' positions $\vec{D'}^{it}$ and their swimming velocity $\vec{V'}^{it}$, as expressed in Eq.~(\ref{eq:explo}).

\begin{equation}\label{eq:explo}
	\vec{D'}^{t}=\vec{D'}^{t-1}+\vec{V'}^{t}
\end{equation}
he velocity is initialized randomly within the range  [$V'_{min}$, $V'_{max}$] and its size is selected based on the size of the problem.

\subsubsection{Exploitation phase}
At this stage, dolphins identify the current optimal solution, which is the nearest prey within the search space. Using the current best positions as references, the other dolphins then adjust their positions toward this optimal solution. This behavior can be mathematically represented as follows:

\begin{equation}\label{eq:update}
	\begin{aligned}
		\vec{A'} & =\left|\vec{C'} \vec{D'}^{* it-1}-\vec{D'}^{it-1}\right| \\
		\vec{D'}^{t} & =\vec{D'}^{* it-1} \cdot \sin (2 \pi l)-\vec{K'} \cdot \vec{A'}
	\end{aligned}
\end{equation}

Where $\mathrm{it}$ represents the current iteration, $\vec{C'}$ and $\vec{K'}$ are coefficient vectors, $\vec{D'}$ signifies the present positions of the dolphins, and $\vec{D'}^{*}$ corresponds to the best achieved dolphin positions. It is worth noting that the leading dolphin follows a circular trajectory, rapidly undulating its tail to create a sinusoidal wave pattern, resulting in the emission of a plume. Meanwhile, the other dolphins encircle the prey in a circular fashion. The computation of the vector $\vec{C}$ is as follows:

Where $\mathrm{it}$ denotes the current iteration, $\vec{C'}$ and $\vec{K'}$ are coefficient vectors, $\vec{D'}$ represents the current positions of the dolphins, and $\vec{D'}^{*}$ corresponds to the best achieved dolphin positions. It is important to note that the leading dolphin follows a circular trajectory, undulating its tail rapidly to generate a sinusoidal wave pattern, which results in the emission of a plume. Concurrently, the other dolphins surround the prey in a circular manner. The computation for the vector $\vec{C}$ is as follows:

\begin{equation}
	\vec{C'} = 2 \cdot \vec{rand}
\end{equation}

By introducing the random vector $\vec{rand}$, we can access any location within the search region. Consequently, Eq.~(\ref{eq:update}) mimics the behavior of dolphins encircling the prey, allowing any dolphin to align its position more closely with the current optimal position.

\subsection{Bubble-net attacking method}

This method draws inspiration from the hunting behavior of whales when searching for prey. It involves a pursuit strategy resembling a spiral pattern, and it is described in Eq.~(\ref{eq:bubble}).

\begin{equation}\label{eq:bubble}
	X(i + 1) = \vec{A'} \cdot e^{bl} \cdot \cos(2\pi l) + \vec{D'}^{* it-1}
\end{equation}
where $b$ is a fixed value set to 1 and is used to determine the configuration of the logarithmic spiral, $l$ is a random value within the range [-1, 1], and the dot (.) represents element-wise multiplication.

\subsection{Cross entropy}
In \cite{13}, the concept of cross entropy is introduced to assess the similarity between two probability distributions. It is formulated in Eq.~(\ref{eq:crossed}) as follows:

\begin{equation}\label{eq:crossed}
	\Gamma(R, S)=\sum_{i=1}^{N} r_{i} \log \left(r_{i} / s_{i}\right)
\end{equation}

In this context, $R=\left\{r_{1}, r_{2}, \ldots, r_{\mathrm{N}}\right\}$ and $S=\left\{s_{1}, s_{2}, \ldots, s_{N}\right\}$ represent the two probability distributions within the same dataset.

By minimizing the cross entropy during image segmentation, the most suitable thresholds are identified and subsequently used to segment the original image. Let $I_{ori}$ be the original image, with its intensity levels represented by $G$. The probability distribution is defined in Eq.~(\ref{eq:nor_hist}).

\begin{equation}
	\label{eq:nor_hist}
	hist^{ch}(i)=\frac{P_{r}^{ch}(i)}{N_{P}}, ch=\left\{\begin{array}{cc}
		1,2,3 & \text { if color image } \\
		1 & \text { if Gray scale image }
	\end{array}\right.
\end{equation}

Here, $i$ represents the intensity levels ranging from 0 to $G-1$. $N_{P}$ denotes the total pixel count within an image. Concurrently, let's assume that the image $I_{ori}$ can be divided into $K$ regions using thresholds $t_{k}^{ch}$, where $k=1,2, \ldots, K-1$. In this context, the cross entropy is expressed as follows:

\begin{equation}\label{eq:cross}
	\resizebox{0.99\linewidth}{!}{$
	\begin{aligned}
		\Gamma^{ch}(t) &= -\sum_{i=1}^{t_{1}-1} i \cdot hist^{ch}(i) \log \left(i / u_{0}^{ch}\right) \
		& -\sum_{i=t_{1}}^{t_{2}-1} i \cdot hist^{ch}(i) \log \left(i / u_{1}^{ch}\right) \
		& -\ldots \\
		& -\sum_{i=t_{k-1}}^{t_{k}-1} i \cdot hist^{ch}(i) \log \left(i / u_{k-1}^{ch}\right) \
		& -\sum_{i=t_{k}}^{L} i \cdot hist^{ch}(i) \log \left(i / u_{k}^{ch}\right)
	\end{aligned}	$}
\end{equation}

where,

 $u_{0}^{ch}=\frac{\sum_{i=1}^{t_{1}-1} i \cdot hist^{ch}(i)}{\sum_{i=1}^{t_{1}-1} i \cdot hist^{ch}(i)}, u_{1}^{ch}=\frac{\sum_{i=t_{1}}^{t_{2}-1} i \cdot hist^{ch}(i)}{\sum_{i=t_{1}}^{t_{2}-1} i \cdot hist^{ch}(i)}, \ldots,\\ \ldots u_{k-1}^{ch}=\frac{\sum_{i=t_{k-2}}^{t_{k-1}-1} i \cdot hist^{ch}(i)}{\sum_{i=t_{k-2}}^{t_{k-1}-1} i \cdot hist^{ch}(i)}, u_{k}^{ch}=\frac{\sum_{i=t_{k}}^{L} i \cdot hist^{ch}(i)}{\sum_{i=t_{k}}^{L} i \cdot hist^{ch}(i)}$.

The MCE determines the optimal threshold $t^{ch *}$ by minimizing the cross entropy as indicated in Eq.~(\ref{eq:min}).

\begin{equation}\label{eq:min}
	t^{ch*}=\arg \min \Gamma^{ch}(t)
\end{equation}
\subsection{The Proposed algorithm}
Algorithm~\ref{alg:mra} presents the pseudocode for the proposed WMRA method, specifically designed for color image segmentation in the context of skin cancer detection. In summary, the method operates as follows:
Initially, it's imperative to set several WMRA parameters. These include the lower bound (lb), upper bound (ub), population size of dolphins ($N$), random population initialization $\vec{D'}^{it}$,  problem dimension ($dim$), maximum number of iterations (${\mathrm{Max} _itr}$), velocity $\vec{V'}$, and the input image retrieval from the skin cancer dataset. Subsequently, the image's normalized histogram is calculated, and the initial fitness values are evaluated using Eqs.~(\ref{eq:nor_hist}) and~(\ref{eq:crossed}), respectively.

\begin{algorithm}
	\caption{The proposed WMRA algorithm pseudo code}
	\label{alg:mra}
	\textbf{Input}: Original color skin cancer image.\\ 
	\textbf{Output}: Image segmented.
	\begin{algorithmic}[1]
		\STATE Initialization of $dim$, $lb$, ,$ub$, $\vec{D'}$, $\vec{V'}$, $N$. 
		\STATE Calculate the normalized histogram using Eq.~(\ref{eq:nor_hist}).
		\STATE Evaluate the objective function of the red, green and blue channels using Eq.~(\ref{eq:cross}).
		\STATE  Obtain the optimal dolphin positions $\vec{D'^*}$
		\WHILE{$(it < {\mathrm{Max} \_itr})$}
		\FOR{$i = 1$ to $dim$}
		\STATE Adjust $K'$, $C'$, $a'$, and $l'$
		\IF{$p < 0.5$}
		\IF{$|K'| \geq 1$}
		\STATE Create fresh solutions by adjusting velocity $v'_i$ through the utilization of Eq.~(\ref{eq:explo}).
		\ELSE
		\STATE \COMMENT{Forming the Mud Ring}
		\STATE Update the Current Dolphin Location Using Eq.~(\ref{eq:update})
		\ENDIF
		\ELSE
		\STATE\COMMENT{Forming the bubble net attack}
		\STATE Update the positions of Dolphin locations using Eq.~(\ref{eq:bubble})
		\ENDIF
		\ENDFOR
		\STATE Check if the positions get outside the bounds
		\STATE Achieve the desired fitness functions of the dolphins.
		\STATE Update the value of $D'^*$ if a best position is present.
		\STATE $it \leftarrow it + 1$
		\ENDWHILE
		\RETURN $D'^*$ (Return the optimal set of thresholds)
	\end{algorithmic}
\end{algorithm}
After determining the best dolphin position, the optimization phase begins. At its core, this phase is distinguished by a key enhancement in this approach: the integration of the bubble-net attack into the foundational MRA algorithm. This change facilitates the updating of dolphin positions with a consistent probability of 0.5, denoted by the variable $p$. This adjustment considerably elevates the solution generation capacity of the modified MRA compared to its original counterpart. For instance, if both $p < 0.5$ and $\vec{K'} < 1$, the dolphins encircle the prey, forming a ring shape as described by Eq.~(\ref{eq:update}). In contrast, when $p$ is greater than or equal to 0.5, the bubble-net attack is initiated, encircling the prey in a spiral formation as given by Eq.~(\ref{eq:bubble}). To conclude, the proposed methodology determines the optimum solution using the best threshold values, as indicated by Eq.~(\ref{eq:min}).

\section{Results and Discussion}\label{sec:Results}
This section presents the results acquired using the WMRA approach in conjunction with the minimum cross-entropy method for segmenting skin color images. The medical images of skin cancer were sourced from a recent dataset.

For comparative purposes, four contemporary optimization techniques were employed: Chimp, RSA, MRA, and WOA. The parameter configurations for these methods adhered to the default values outlined in their respective original publications. To ensure an unbiased comparison, each method was executed independently 15 times. Furthermore, the dolphin population $N$ was fixed at 25, and the maximum iteration count $Max_itr$ was capped at 300.

\begin{figure}[t!]
	\centering
	\includegraphics[width=0.7\linewidth]{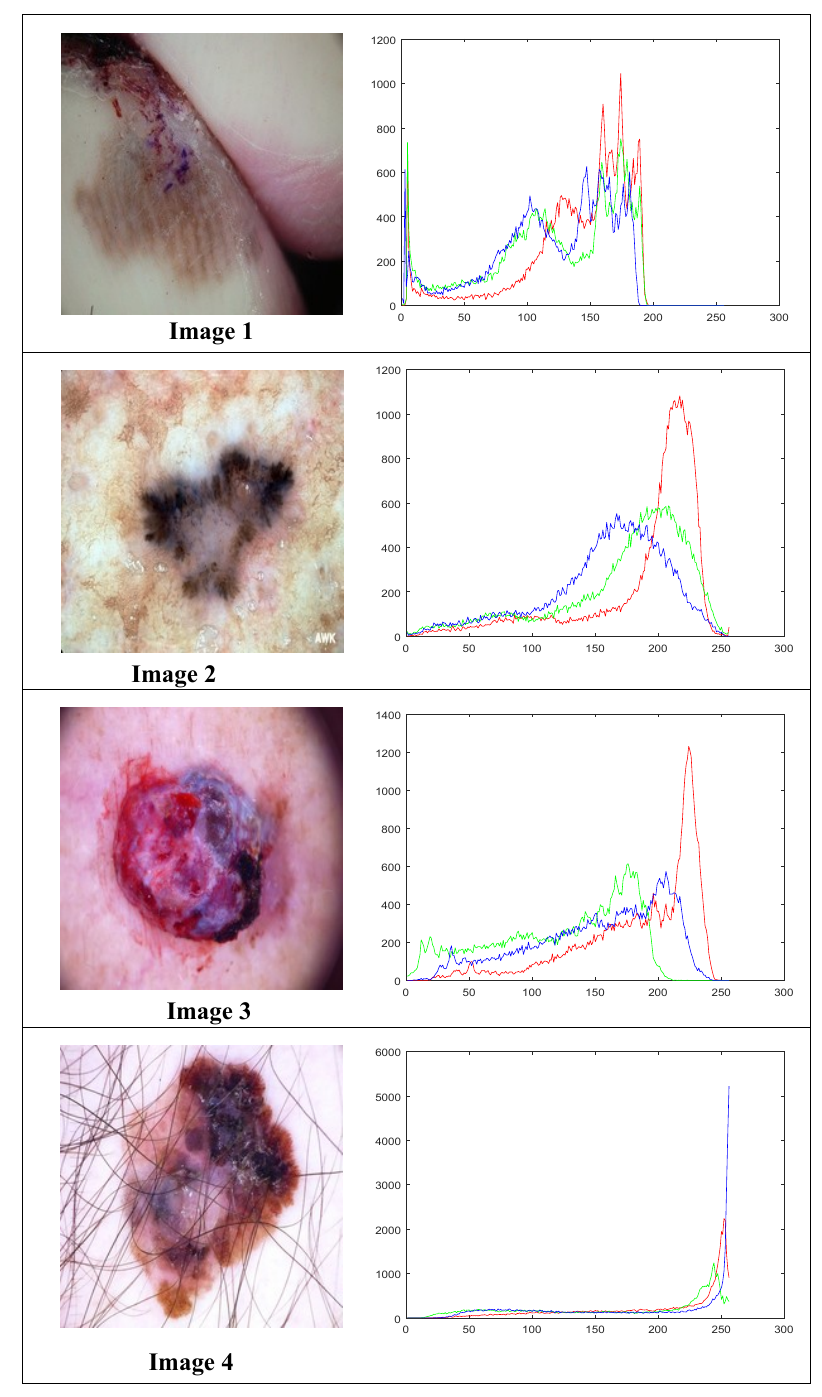}
	\caption{Skin cancer images for various patients.}
	\label{fig:dataset}
\end{figure}

The segmentation evaluation spanned various threshold values (nTh = 2, 3, 4, and 5) and encompassed three assessment metrics: fitness (across the red, green, and blue channels), Peak Signal to Noise Ratio (PSNR), and Mean Square Error (MSE)~\cite{14}. Additionally, Fig.~\ref{fig:dataset} showcases the images and their associated histograms utilized during the segmentation process.

\begin{figure}[t!]
	\centering
	\includegraphics[width=0.7\linewidth]{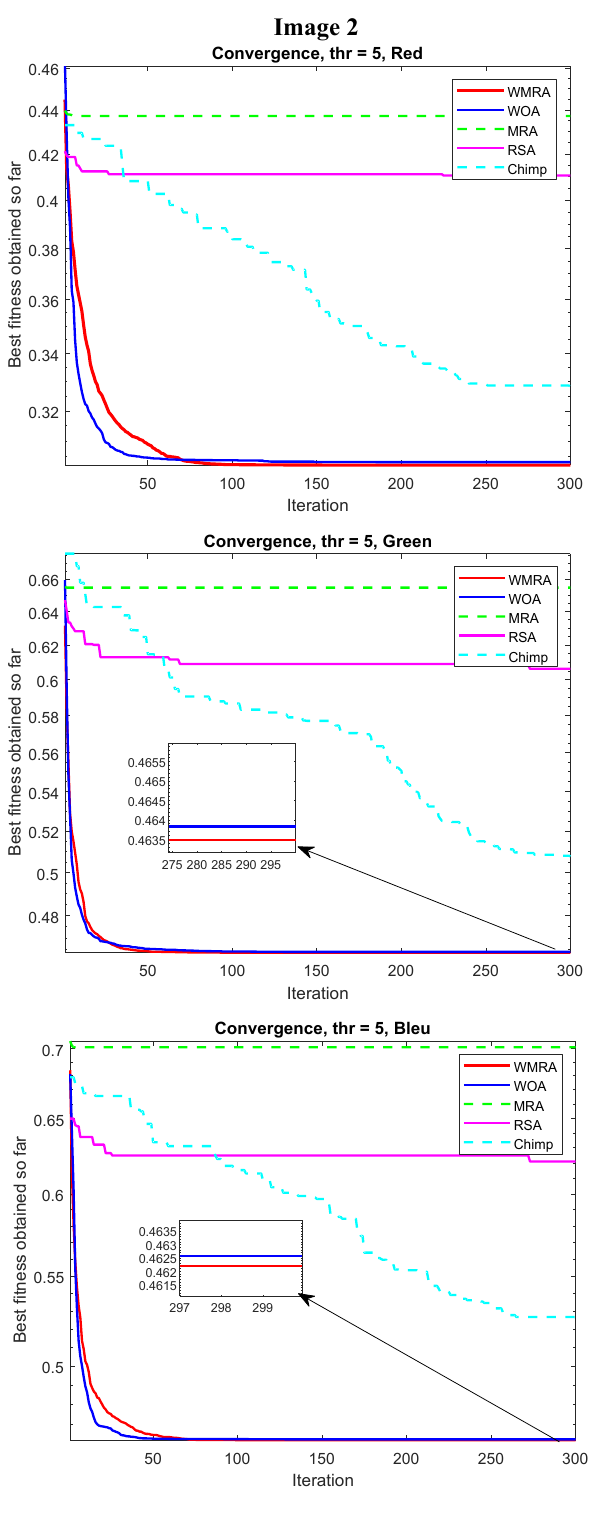}
	\caption{Image 2 convergence curves at level 5.}
	\label{fig:convergence}
\end{figure}

\begin{figure}[t!]
	\centering
	\includegraphics[width=0.95\linewidth]{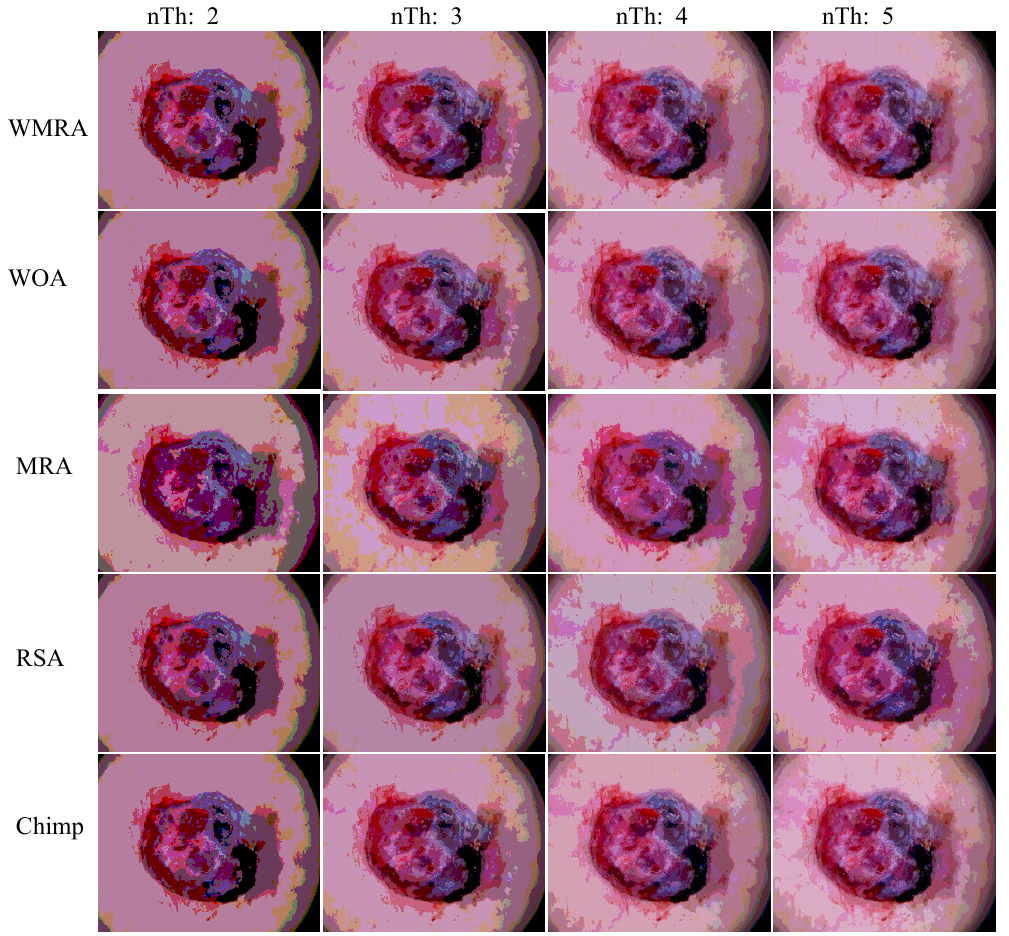}
	\caption{Samples of the segmented image 3 at all thresholds.}
	\label{fig:img_seg}
\end{figure}

Tab.~\ref{tab:red}, Tab.~\ref{tab:green}, and Tab.~\ref{tab:blue} display the average fitness results for the red, green, and blue channels, respectively. It's noteworthy that lower fitness values highlight the efficacy of the proposed methods in searching for optimal solutions and adeptly avoiding local optima. As per the results, the proposed method yields the most favorable outcomes when compared to its counterparts, with the WOA algorithm trailing closely in terms of fitness across all channels (red, green, and blue). For a more comprehensive analysis, the convergence curve for the threshold value of 5 for Image 2 is depicted in Fig.~\ref{fig:convergence}. Additionally, Fig.~\ref{fig:img_seg} showcases the qualitative outcomes of the segmented image, labeled as Img 3, across all threshold levels.

\begin{table}[t!]
\centering
\caption{Comparison of fitness of Red channel values obtained by WMRA, WOA, MRA, Chimp, RSA using minimum cross entropy.}

	\label{tab:red}
	\resizebox{0.95\linewidth}{!}{
	\begin{tabular}{ccccccc}
	\hline
		Image & nTh &WMRA & WOA & MRA & Chimp & RSA\\
	\hline
	Img 1 & 2 & \textbf{1.3171} & \textbf{1.3171} & 1.5071 & 1.3610 & 1.3172\\
	& 3 & \textbf{0.7683} & 0.7683 & 1.0319 & 0.9168 & 0.7709\\
	& 4 & \textbf{0.5182} & 0.5186 & 0.7631 & 0.6554 & 0.5478\\
	& 5 & \textbf{0.3551} & 0.3584 & 0.5854 & 0.5347 & 0.3855\\
	Img 2 & 2 & \textbf{1.0522} & \textbf{1.0522} & 1.1466 & 1.0830 & 1.0523\\
	& 3 & \textbf{0.6406} & 0.6407 & 0.7251 & 0.7340 & 0.6433\\
	& 4 & \textbf{0.4279} & 0.4280 & 0.5509 & 0.5510 & 0.4350\\
	& 5 & \textbf{0.3021} & 0.3031 & 0.4372 & 0.4105 & 0.3284\\
	Img 3 & 2 & \textbf{1.1094} & 1.1094 & 1.1865 & 1.1560 & 1.1096\\
	& 3 & \textbf{0.5865} & 0.5866 & 0.7741 & 0.7188 & 0.6269\\
	& 4 & \textbf{0.3833} & 0.3834 & 0.5557 & 0.5469 & 0.4219\\
	& 5 & \textbf{0.2758} & 0.2761 & 0.4343 & 0.4286 & 0.3340\\
	Img 4 & 2 & \textbf{1.1354} & 1.1354 & 1.2173 & 1.1787 & 1.1357\\
	& 3 & \textbf{0.6523} & \textbf{0.6523} & 0.7812 & 0.7969 & 0.6566\\
	& 4 & \textbf{0.4162} & 0.4163 & 0.5736 & 0.5771 & 0.4291\\
	& 5 & \textbf{0.2893} & 0.2893 & 0.4370 & 0.4387 & 0.3189\\
	\hline
	\end{tabular}}
\end{table}

\begin{table}[h!]
	\centering
	\caption{Comparison of fitness of Green channel values obtained by WMRA, WOA, MRA, Chimp, RSA using minimum cross entropy.}
	\label{tab:green}
	\resizebox{0.95\linewidth}{!}{
		\begin{tabular}{ccccccc}
			\hline
			Image & nTh &WMRA & WOA & MRA & Chimp & RSA\\
			\midrule
			Img 1 & 2 & \textbf{1.7981} & 1.7981 & 2.0902 & 1.8770 & 1.7984\\
			& 3 & \textbf{1.0072} & 1.0073 & 1.4602 & 1.1679 & 1.0131\\
			& 4 & \textbf{0.6513} & 0.6516 & 0.9966 & 0.8836 & 0.6666\\
			& 5 & \textbf{0.4664} & 0.4669 & 0.8527 & 0.6736 & 0.5198\\
			Img 2 & 2 & \textbf{1.6249} & \textbf{1.6249} & 1.7312 & 1.7257 & 1.6253\\
			& 3 & \textbf{1.0286} & 1.0286 & 1.1969 & 1.1334 & 1.0323\\
			& 4 & \textbf{0.6360} & 0.6362 & 0.8553 & 0.7767 & 0.6625\\
			& 5 & \textbf{0.4635} & 0.4638 & 0.6546 & 0.6063 & 0.5078\\
			Img 3 & 2 & \textbf{1.8206} & \textbf{1.8206} & 2.0744 & 1.8613 & 1.8210\\
			& 3 & \textbf{1.0803} & 1.0804 & 1.4834 & 1.2727 & 1.0852\\
			& 4 & \textbf{0.6830} & 0.6834 & 1.0543 & 0.8523 & 0.7030\\
			& 5 & \textbf{0.4989} & 0.5000 & 0.7804 & 0.7491 & 0.5503\\
			Img 4 & 2 & \textbf{1.7608} & 1.7608 & 1.8546 & 1.8161 & 1.7611\\
			& 3 & \textbf{0.9851} & 0.9852 & 1.2269 & 1.1179 & 0.9920\\
			& 4 & \textbf{0.6667} & 0.6668 & 0.9016 & 0.8259 & 0.6911\\
			& 5 & \textbf{0.4647} & 0.4648 & 0.6368 & 0.6597 & 0.5142\\
			\hline
	\end{tabular}}
\end{table}

\begin{table}[h!]
	\centering
	\caption{Comparison of fitness of Bleu channel values obtained by WMRA, WOA, MRA, Chimp, RSA using minimum cross entropy.}
	\label{tab:blue}
	\resizebox{0.95\linewidth}{!}{
		\begin{tabular}{ccccccc}
			\toprule
						Image & nTh &WMRA & WOA & MRA & Chimp & RSA\\
						\midrule
Img 1 & 2 & \textbf{1.711232} & 1.711238 & 2.0835 & 1.7596 & 1.7114\\
& 3 & \textbf{0.9878} & 0.9881 & 1.3227 & 1.2098 & 0.9926\\
& 4 & \textbf{0.6864} & 0.6865 & 1.0626 & 0.8341 & 0.7164\\
& 5 & 0.4895 & \textbf{0.4608} & 0.7895 & 0.6933 & 0.5123\\
Img 2 & 2 & \textbf{1.6034} & \textbf{1.6034} & 1.7324 & 1.6505 & 1.6039\\
& 3 & \textbf{0.9848} & 0.9848 & 1.1856 & 1.1466 & 0.9911\\
& 4 & \textbf{0.6462} & 0.6462 & 0.9201 & 0.8621 & 0.6666\\
& 5 & \textbf{0.4622} & 0.4626 & 0.7008 & 0.6210 & 0.5265\\
Img 3 & 2 & \textbf{1.4891} & \textbf{1.4891} & 1.6011 & 1.5421 & 1.4897\\
& 3 & \textbf{0.8554} & 0.8554 & 1.1252 & 1.0171 & 0.8647\\
& 4 & \textbf{0.5536} & 0.5537 & 0.8146 & 0.7842 & 0.6154\\
& 5 & \textbf{0.3908} & 0.3909 & 0.6258 & 0.5862 & 0.4282\\
Img 4 & 2 & \textbf{1.378669} & 1.378691 & 1.5351 & 1.4218 & 1.3793\\
& 3 & \textbf{0.7608} & 0.7608 & 0.9068 & 0.8748 & 0.7689\\
& 4 & \textbf{0.4882} & 0.4882 & 0.6929 & 0.5798 & 0.5030\\
& 5 & \textbf{0.3402} & 0.3503 & 0.5131 & 0.4942 & 0.3784\\
			\bottomrule
	\end{tabular}}
\end{table}

Tab.~\ref{tab:psnr_MCE} displays the average values procured by the WMRA across various threshold levels (2, 3, 4, and 5). High PSNR values are indicative of successful segmentation results. Notably, the WMRA method consistently produces elevated PSNR values for both Img 3 and Img 4 across all thresholds. Furthermore, it yields promising results for Img 2 at thresholds 2, 4, and 5. For Img 1, the method is particularly effective at levels 2 and 4. In a similar vein, the RSA approach matches the PSNR results for images labeled as Img 1 and Img 2 at threshold 3. Subsequently, the WOA algorithm registers a PSNR of 22.836 for Img 1 at threshold 5, showcasing its competitive performance.

\begin{table}[h!]
	\centering
	\caption{Comparison mean PSNR values obtained by WMRA, WOA, MRA, Chimp, RSA using minimum cross entropy.}
	\label{tab:psnr_MCE}
	\resizebox{0.95\linewidth}{!}{
		\begin{tabular}{ccccccc}
			\toprule
			Image & nTh &WMRA & WOA & MRA & Chimp & RSA\\
			\midrule
Img 1 & 2 & \textbf{15.471} & 15.448 & 15.109 & 15.024 & 15.453\\
& 3 & 18.670 & 18.625 & 17.391 & 16.778 & \textbf{18.709}\\
& 4 & \textbf{20.855} & 20.813 & 18.516 & 18.511 & 20.713\\
& 5 & 22.784 & \textbf{22.836} & 20.038 & 19.646 & 22.552\\
Img 2 & 2 & \textbf{15.135} & 15.127 & 14.811 & 14.517 & 15.134\\
& 3 & 17.858 & 17.853 & 16.815 & 16.347 & \textbf{17.917}\\
& 4 & \textbf{19.992} & 19.992 & 18.294 & 18.224 & 19.914\\
& 5 & \textbf{21.666} & 21.664 & 19.565 & 19.657 & 21.243\\
Img 3 & 2 & \textbf{15.361} & 15.344 & 15.005 & 14.915 & 15.349\\
& 3 & \textbf{18.200} & 18.182 & 16.686 & 16.773 & 18.069\\
& 4 & \textbf{20.304} & 20.274 & 18.458 & 18.522 & 19.982\\
& 5 & \textbf{21.876} & 21.864 & 19.768 & 19.425 & 21.454\\
Img 4 & 2 & \textbf{13.257} & 13.229 & 13.017 & 12.842 & 13.239\\
& 3 & \textbf{16.070} & 16.038 & 15.314 & 14.865 & 15.997\\
& 4 & \textbf{18.149} & 18.134 & 16.864 & 16.543 & 18.036\\
& 5 & \textbf{20.067} & 19.978 & 18.346 & 17.683 & 19.836\\
			\bottomrule
	\end{tabular}}
\end{table}

Tab.~\ref{tab:mse_MCE} provides a comparison of the average MSE values yielded by the WMRA, WOA, Chimp, RSA, and MRA methods. The data indicates that WMRA consistently records low MSE values for Img 3 and Img 4 across all thresholds (2, 3, 4, and 5). Specifically for Img 1, the method posts reduced MSE values at thresholds 2 and 4, though this is not the case for other thresholds. For Img 2, WMRA delivers commendable results at thresholds 2, 4, and 5. On the other hand, both RSA and WOA display positive results for Img 1 and Img 2, with each method excelling at distinct thresholds. RSA boasts impressive performance in two instances, while WOA shines in one.

\begin{table}[h!]
	\centering
	\caption{Comparison mean MSE values obtained by WMRA, WOA, MRA, Chimp, RSA using minimum cross entropy.}
	\label{tab:mse_MCE}
	\resizebox{0.95\linewidth}{!}{
		\begin{tabular}{ccccccc}
			\toprule
			Image & nTh &WMRA & WOA & MRA & Chimp & RSA\\
			\midrule
			Img 1 & 2 & \textbf{1844.863} & 1854.845 & 2037.171 & 2046.934 & 1852.748\\
			& 3 & 883.184 & 892.517 & 1191.664 & 1379.535 & \textbf{875.619}\\
			& 4 & \textbf{534.066} & 539.293 & 931.622 & 924.151 & 553.251\\
			& 5 & 343.018 & \textbf{338.532} & 649.423 & 718.968 & 363.506\\
			Img 2 & 2 & \textbf{1993.146} & 1997.224 & 2156.463 & 2311.868 & 1994.011\\
			& 3 & 1064.851 & 1066.119 & 1375.317 & 1513.198 & \textbf{1050.611}\\
			& 4 & \textbf{651.392} & 651.528 & 969.147 & 980.757 & 663.717\\
			& 5 & \textbf{443.109} & 443.283 & 732.056 & 709.508 & 490.207\\
			Img 3 & 2 & \textbf{1892.235} & 1899.736 & 2058.321 & 2098.847 & 1897.345\\
			& 3 & \textbf{984.145} & 988.305 & 1405.424 & 1374.519 & 1016.817\\
			& 4 & \textbf{606.245} & 610.476 & 933.453 & 923.486 & 656.175\\
			& 5 & \textbf{422.154} & 423.352 & 690.457 & 745.263 & 467.957\\
			Img 4 & 2 & \textbf{3071.522} & 3091.757 & 3255.072 & 3383.091 & 3084.912\\
			& 3 & \textbf{1607.111} & 1619.155 & 1935.182 & 2129.434 & 1635.251\\
			& 4 & \textbf{995.898} & 999.287 & 1349.277 & 1451.544 & 1023.197\\
			& 5 & \textbf{640.240} & 654.187 & 959.267 & 1125.745 & 676.495\\
			\bottomrule
	\end{tabular}}
\end{table}

\section{Conclusion}\label{sec:conclusion}
Skin cancer is one of the most common forms of cancer, rendering early detection critical in significantly reducing the related mortality rates. Image segmentation, utilizing multilevel thresholding, serves as an essential phase in extracting regions of interest from skin cancer images. This study delves into the issue of multilevel thresholding, employing skin cancer images and minimum cross-entropy as a fitness function. We introduce an enhanced version of the Mud Ring Algorithm (MRA), dubbed the Tiled WMRA. This novel method marries the bubble-net attack strategy from the Whale Optimization Algorithm with the mud ring mechanism, constructing a resilient search space. The aim is to pinpoint optimal thresholds and secure precise segmentation results. We subjected our proposed approach to rigorous testing using various metrics, such as fitness, PSNR, and MSE. It was juxtaposed against contemporary algorithms like MRA, Chimp, WOA, and RSA. Based on the results, our WMRA model surpassed its counterparts across all evaluation parameters. For future pursuits, we plan to delve deeper by incorporating a wider range of evaluation metrics and expanding the dataset. Additionally, the potential enhancements from integrating concepts like chaotic theory will be explored.


\end{document}